\documentclass[conference]{IEEEtran}
\IEEEoverridecommandlockouts
\usepackage{cite}
\usepackage{amsmath,amssymb,amsfonts}
\usepackage{algorithmic}
\usepackage{graphicx}
\usepackage{textcomp}
\usepackage{xcolor}
\usepackage{algorithm}

\def\BibTeX{{\rm B\kern-.05em{\sc i\kern-.025em b}\kern-.08em
    T\kern-.1667em\lower.7ex\hbox{E}\kern-.125emX}}
\begin{document}

\title{CAD: Confidence-Aware Adaptive Displacement for Semi-Supervised Medical Image Segmentation\\

\thanks{This work was funded by the Scientific Research Key project of Education Office of Hunan Province, China [24A0176] \& the National Key Research and Development Program [2021YFD1300404].}
}

\author{
    \IEEEauthorblockN{Wenbo Xiao$^{a,b}$, Zhihao Xu$^{a}$, Guiping Liang$^{a}$, Yangjun Deng$^{a}$, Yi Xiao$^{a*}$\thanks{* Yi Xiao is corresponding author.}}
    \IEEEauthorblockA{$^a$College of Information and Intelligence, Hunan Agricultural University, Changsha, China}
    
    \IEEEauthorblockA{$^b$School of Computer Science and Engineering, University of New South Wales, Sydney, Australia}
    \IEEEauthorblockA{\{wenboxiao@stu., guapideyouxiang@stu., guipingliang@stu., dengyangjun@, xiaoyi@\}hunau.edu.cn}
}
\maketitle


\begin{abstract}
Semi-supervised medical image segmentation aims to leverage minimal expert annotations, yet remains confronted by challenges in maintaining high-quality consistency learning. Excessive perturbations can degrade alignment and hinder precise decision boundaries, especially in regions with uncertain predictions. In this paper, we introduce Confidence-Aware Adaptive Displacement (CAD), a framework that selectively identifies and replaces the largest low-confidence regions with high-confidence patches. By dynamically adjusting both the maximum allowable replacement size and the confidence threshold throughout training, CAD progressively refines the segmentation quality without overwhelming the learning process. Experimental results on public medical datasets demonstrate that CAD effectively enhances segmentation quality, establishing new state-of-the-art accuracy in this field.
\end{abstract}


\section{Introduction}

Medical image segmentation is an essential task in many clinical applications, such as disease diagnosis, treatment planning, and surgical guidance. Accurate segmentation allows clinicians to delineate structures such as organs and tumors from imaging modalities like computer tomography (CT) and magnetic resonance imaging (MRI) scans, which is crucial for providing reliable volumetric and shape information\cite{b1}. However, obtaining large datasets with precise annotations is challenging and costly, particularly in the medical imaging domain, where only experts can provide reliable annotations. This creates a significant barrier to training robust deep learning models for segmentation tasks, especially in medical applications where high-quality labeled data is sparse.

In medical image segmentation, semi-supervised medical image segmentation (SSMIS) methods have emerged to enhance model performance by utilizing a small amount of labeled data and a large amount of unlabeled data\cite{b3,b4}. This approach alleviates the challenge of limited labeled data and effectively leverages the rich information contained in the unlabeled dataset. In recent years, as this field has advanced, the performance of SSMIS has increasingly approached that of fully supervised segmentation methods, particularly under conditions with minimal labeled data, showing remarkable segmentation results\cite{b5,b6}. This progress has been driven by several techniques, with consistency learning being one of the core methods in semi-supervised learning\cite{b7,b8,b9}, which facilitates the effective use of unlabeled data by enforcing consistent predictions across different augmentations of the input data, utilizing pseudo-labels and data augmentation strategies to generate high-quality labels\cite{b10}. Common techniques in consistency learning include self-training and consistency loss, which minimize the discrepancy between model outputs under different augmentations, making the model more stable and reliable when processing unlabeled data\cite{b11,b12}. Within consistency learning, perturbations such as data augmentations and feature map perturbations play a pivotal role\cite{b12}, ensuring that the model learns robust features by making predictions consistent despite variations in the input data. Data augmentations, including rotation, scaling, and flipping, challenge the model, forcing it to focus on stable and generalizable features while ignoring spurious ones\cite{b13,b14}; meanwhile, feature map perturbations, such as adding noise or applying dropout, directly alter the feature representations, encouraging the model to maintain consistency even in the presence of noisy or incomplete information\cite{b10,b16}. Together, these perturbations help the model refine its internal representations, ultimately enhancing its performance on unlabeled data and improving segmentation accuracy in semi-supervised settings.

In SSMIS, a key challenge is the balance between applying sufficient perturbations and avoiding excessive ones. Too strong perturbations can increase uncertainty in the model's predictions, which undermines its ability to learn stable representations\cite{b33}, especially in scenarios with limited labeled data. Additionally, in continuous learning settings, the model's ability to extract meaningful semantic information is constrained by the mismatch between labeled and unlabeled data distributions. The small labeled set often does not fully capture the variations in the unlabeled data, limiting the model’s generalization capacity and potentially leading to error propagation over time.

\begin{figure}[htbp]
\centerline{\includegraphics[width=250pt]{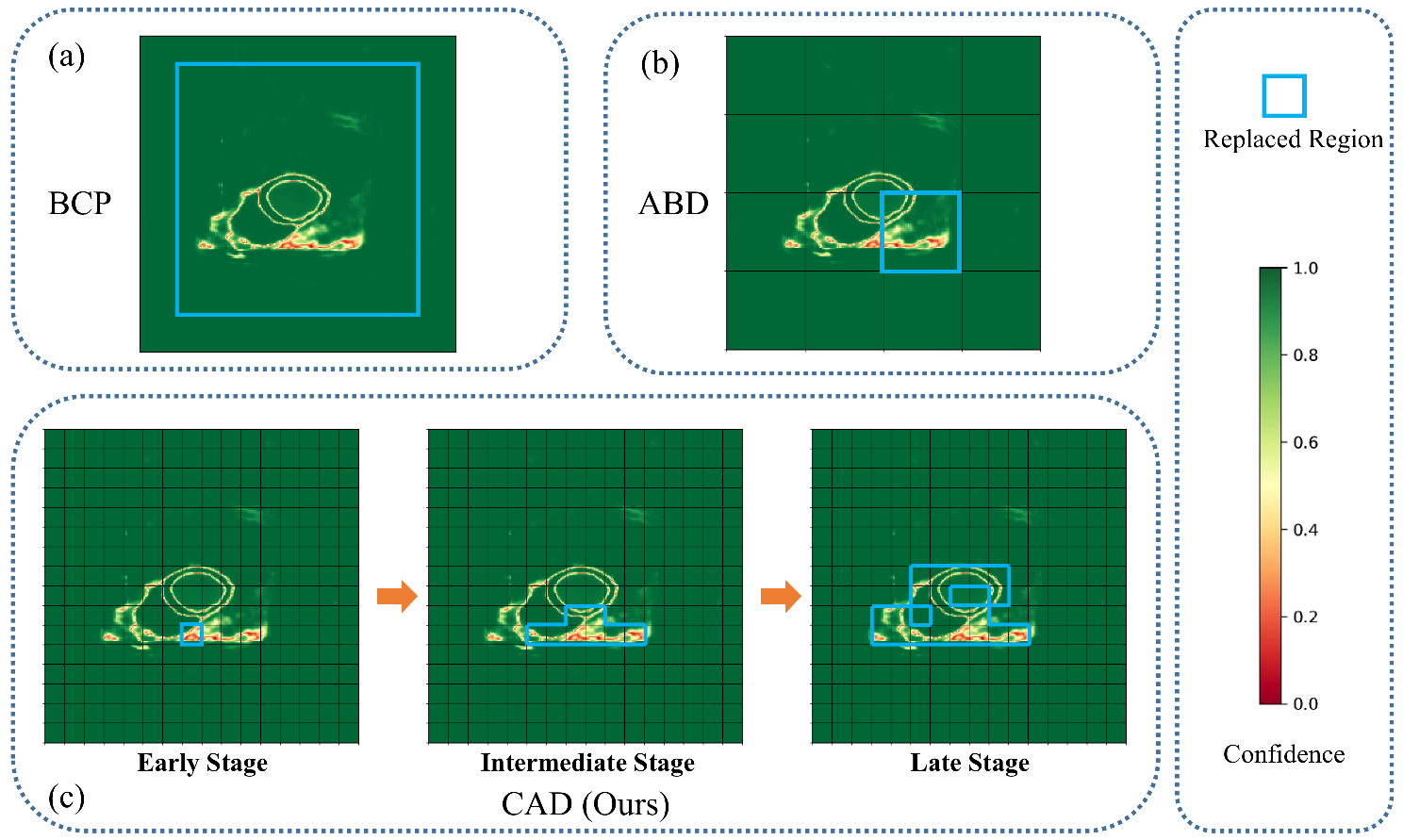}}
\caption{Visualization of replacement  regions in BCP, ABD, and CAD. Subfigure (a) shows BCP, which swaps foreground and background between labeled and unlabeled data across the entire image. Subfigure (b) illustrates ABD, replacing low-confidence patches with high-confidence patches between strong and weak augmentations. Subfigure (c) demonstrates the proposed CAD, progressively refining replacement regions from small (early stage) to large (late stage) based on confidence levels and training dynamics.}
\label{fig1}
\end{figure}

Recent research has made significant strides in addressing these challenges. As shown in Fig.~\ref{fig1}(a), Bidirectional Copy-Paste (BCP)\cite{b3} tackles the distribution mismatch by swapping foreground and background patches between labeled and unlabeled images. This strategy encourages the model to learn from mixed pseudo-labels, enabling better generalization. Adaptive Bidirectional Displacement (ABD)\cite{b5}, depicted in Fig.~\ref{fig1}(b), replaces low-confidence regions in weakly augmented images with high-confidence regions from strongly augmented counterparts and vice versa, thus refining consistency learning by targeted perturbations. However, both approaches have limitations. The reliance on mixed pseudo-labels in BCP can propagate errors from unreliable predictions, while ABD's rigid patch-replacement mechanism may overlook nuanced contextual dependencies, limiting its adaptability across varying data complexities.

This work builds on the insights from ABD and addresses its limitations with a novel Confidence-Aware Displacement (CAD) strategy. As illustrated in Fig.~\ref{fig1}(c), CAD dynamically escalates the replacement process across training stages. In the early stages, the replacement focuses on small, low-confidence regions, ensuring stable learning. As training progresses, CAD adaptively increases the scale of replacement by refining both the confidence threshold and the region size. This progressive approach not only mitigates the impact of unstable regions but also ensures that the model effectively integrates reliable information from both labeled and unlabeled data. By introducing this adaptive mechanism, CAD enhances segmentation performance and provides a more robust framework for handling challenging semi-supervised scenarios.

Our main contributions can be summarized as follows:
\begin{itemize}
    \item We propose a novel CAD strategy that incorporates the Largest Low-Confidence Region Replacement (LLCR) method to dynamically replace low-confidence regions with high-confidence counterparts during training, enabling more stable and robust learning.
    \item We introduce a Dynamic Threshold Escalation (DTE) mechanism, which adaptively adjusts the confidence threshold and replacement region size throughout training, ensuring that the replacement process aligns with the evolving model predictions.
    \item Extensive experiments on public datasets demonstrate that our CAD framework achieves state-of-the-art performance, significantly surpassing existing methods in both accuracy and robustness.
\end{itemize}

\section{Related Works}

Consistency learning plays a pivotal role in SSMIS by effectively utilizing large amounts of unlabeled data. The core idea is to encourage models to produce stable predictions under various perturbations or augmentations of the input data, thereby enhancing performance even with limited labeled examples. Early works, such as the study by Sajjadi et al. \cite{b7}, highlighted the importance of stochastic transformations like dropout and random augmentations to regularize deep networks, improving generalization and robustness against input variability. More recent studies have expanded on these ideas to enhance consistency in semi-supervised segmentation tasks; for instance, ConMatch\cite{b31} incorporates confidence-guided consistency regularization, refining pseudo-label confidence estimations to improve performance. Additionally, methods such as UDiCT\cite{b32} pair annotated and unannotated data based on uncertainty, thereby mitigating the impact of unreliable pseudo-labels, while Huang et al. \cite{b33} introduced a two-stage approach enforcing consistency across perturbed versions of unlabeled electron microscopy volumes to enhance model robustness.

In medical image segmentation, consistency learning addresses significant challenges posed by data variability and insufficient labeled data. Prominent methods have augmented consistency learning by integrating additional tasks or uncertainty estimations, as demonstrated by Shu et al. \cite{b34} and Wang et al. \cite{b35}. For instance, Liu et al. \cite{b36} applied transformer-based models to COVID-19 lesion segmentation, using consistency across augmented views to alleviate the shortage of labeled data. Moreover, Chen et al. \cite{b37} proposed Cross Pseudo Supervision (CPS), where two segmentation networks with different initializations enforce consistency on each other's predictions, effectively expanding training data via pseudo-labels. Another notable contribution is from Tarvainen and Valpola \cite{b9}, who introduced the Mean Teacher model, maintaining an exponential moving average (EMA) of model weights to improve segmentation performance under semi-supervised conditions. Collectively, these advances demonstrate how consistency learning, especially when combined with innovative model architectures and training strategies, significantly enhances segmentation performance with limited labeled data.

\section{Method}

Mathematically, we define the 3D volume of a medical image as $X \in \mathbb{R}^{W \times H \times L}$, where $W$, $H$, and $L$ denote the width, height, and depth of the volume, respectively. The goal of semi-supervised medical image segmentation is to predict the per-voxel label map $Y \in \{0,1,\dots,K-1\}^{W \times H \times L}$, indicating where the background and the targets are in $X$. Here, $K$ represents the number of classes. The training set $\mathcal{D}$ consists of $N$ labeled data and $M$ unlabeled data ($N \ll M$), expressed as two subsets: $\mathcal{D} = \mathcal{D}_l \cup \mathcal{D}_u$, where $\mathcal{D}_l = \{(X^l_i, Y^l_i)\}^N_{i=1}$ and $\mathcal{D}_u = \{X^u_i\}^{M+N}_{i=N+1}$. In subsequent descriptions, we omit the index $i$ for simplicity.




\begin{figure*}[htbp]
\centerline{\includegraphics[width=520pt]{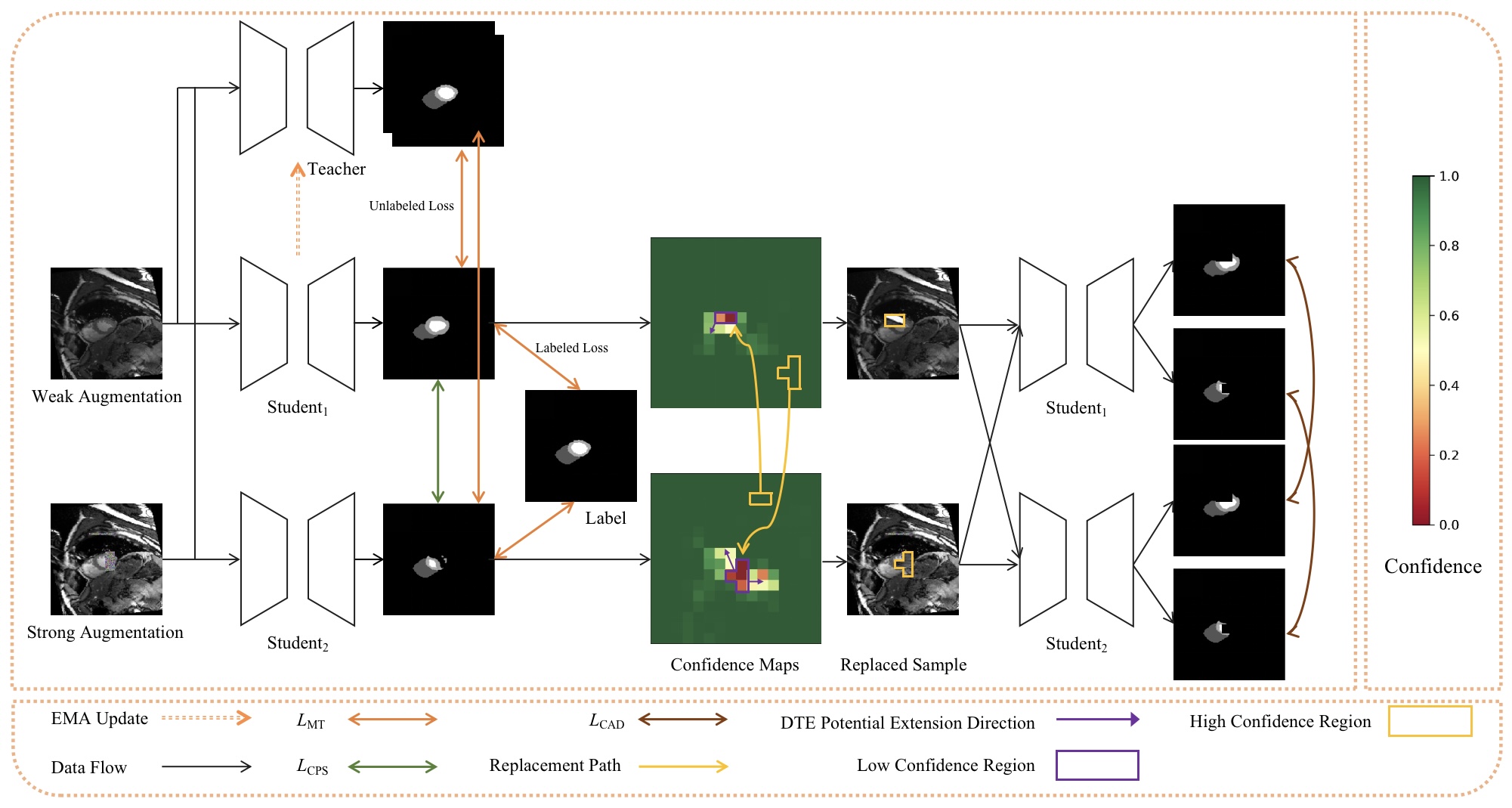}}
\caption{Overview of the proposed framework integrating Mean Teacher with DTE, LLCR, and CPS. The diagram illustrates the data flow, confidence-based region replacement, and collaborative learning between the student and teacher models.}
\label{fig2}
\end{figure*}

The overall workflow of our proposed method is as follows:

\begin{enumerate}
    \item Mean Teacher Framework: The teacher model \(f_t\) is updated via EMA of student models \(f_{\theta_1}\) and \(f_{\theta_2}\). These student models process weakly (\(X_w\)) and strongly (\(X_s\)) augmented images, respectively.

    \item Dynamic Threshold Escalation (DTE) and Largest Low-Confidence Region Replacement (LLCR): In each iteration, confidence maps derived from \(f_{\theta_1}\) and \(f_{\theta_2}\) guide the identification of the largest low-confidence region (\(\mathcal{R}_\mathrm{low}\)) in \(X_s\). This region is replaced by a high-confidence counterpart from \(X_w\), and vice versa, ensuring the model focuses on uncertain areas while maintaining robust learning dynamics.

    \item Cross Pseudo Supervision (CPS): After LLCR, predictions from both models are compared using CPS loss\cite{b37}, which further refines the segmentation by enforcing consistency between the two student models.

\end{enumerate}

The combined strategies, as illustrated in Fig.~\ref{fig2}, enable our framework to address uncertainty effectively and improve semi-supervised segmentation performance.

\subsection{Confidence-Aware Adaptive Displacement}
\subsubsection{Largest Low-Confidence Region Replacement (LLCR)}

In the segmentation process, the input image is divided into smaller patches. Specifically, we partition the image into $h \times w$ patches, resulting in $N_p$ total patches, where $N_p = h \times w$. Let $p_i$ denote the $i$-th patch, with $i \in \{1, \dots, N_p\}$. Each patch represents a localized region of the image, which is crucial for segmenting and replacing low-confidence areas at the patch level.

Let $f_{\theta_1}$ and $f_{\theta_2}$ denote the two student models. For an input image $X_w$ with weak augmentation and $X_s$ with strong augmentation, the models produce logits as follows:
\begin{equation}
\text{logits}_w = f_{\theta_1}(X_w), \quad \text{logits}_s = f_{\theta_2}(X_s). \label{eq:logits}
\end{equation}

Applying the softmax operation to the logits, we obtain the per-pixel probabilities:
\begin{equation}
\begin{aligned}
P_w(x) &= \mathrm{softmax}(\text{logits}_w(x)), \\
P_s(x) &= \mathrm{softmax}(\text{logits}_s(x)), 
\end{aligned}
\quad
\forall x \in \Omega,
\label{eq:softmax}
\end{equation}

where $\Omega$ denotes the spatial domain of the image. For each pixel $x$, the confidence value is defined as the maximum probability across all classes:
\begin{equation}
\alpha_w(x) = \max_{c \in \{1, \dots, K\}} P_{w,c}(x), 
\quad 
\alpha_s(x) = \max_{c \in \{1, \dots, K\}} P_{s,c}(x). 
\label{eq:confidence}
\end{equation}

The confidence of a patch $p_i$ is computed as the average confidence of all pixels within the patch:
\begin{equation}
C_{p_i} = \frac{1}{|p_i|} \sum_{x \in p_i} \alpha(x),
\label{eq:patch_confidence}
\end{equation}
where $|p_i|$ is the total number of pixels in patch $p_i$.

To standardize the confidence values, we apply min-max normalization:
\begin{equation}
\widetilde{C}_{p_i} = \frac{C_{p_i} - \min_j C_{p_j}}{\max_j C_{p_j} - \min_j C_{p_j}},
\label{eq:minmax_normalization}
\end{equation}
where $\widetilde{C}_{p_i} \in [0, 1]$ ensures that confidence values are scaled to a consistent range for comparison.

To identify the largest low-confidence region, we use a Breadth-First Search (BFS) approach starting from the patch with the lowest confidence. The BFS process is guided by two thresholds:
- $C_\mathrm{threshold}$: the confidence threshold for including patches.
- $R_\mathrm{threshold}$: the maximum size of the connected region.

The BFS algorithm is described using pseudocode in the following format:

\begin{algorithm}[H]
\caption{BFS for Largest Low-Confidence Region}
\begin{algorithmic}[1]
\STATE \textbf{Initialize:} $\mathcal{R}_\mathrm{low} \gets \emptyset$
\STATE Identify the patch $p_0$ with the lowest confidence:
\[
p_0 = \arg\min_i \widetilde{C}_{p_i}.
\]
\STATE Push $p_0$ into a priority queue (min-heap) keyed by $\widetilde{C}_{p_i}$.
\WHILE {the priority queue is not empty and $|\mathcal{R}_\mathrm{low}| < R_\mathrm{threshold}$}
    \STATE Pop the patch $p$ with the lowest confidence from the queue.
    \IF {$p \notin \mathcal{R}_\mathrm{low}$ and $\widetilde{C}_{p} \leq C_\mathrm{threshold}$}
        \STATE Add $p$ to $\mathcal{R}_\mathrm{low}$.
    \ENDIF
    \FORALL {neighbors $q$ of $p$}
        \IF {$q \notin \mathcal{R}_\mathrm{low}$ and $\widetilde{C}_{q} \leq C_\mathrm{threshold}$}
            \STATE Push $q$ into the queue.
        \ENDIF
    \ENDFOR
\ENDWHILE
\STATE \textbf{Output:} $\mathcal{R}_\mathrm{low}$ as the largest low-confidence region.
\end{algorithmic}
\end{algorithm}

The resulting connected region $\mathcal{R}_\mathrm{low}$ is described using shape offsets. Let the bounding box of $\mathcal{R}_\mathrm{low}$ extend from $(r_\mathrm{min}, c_\mathrm{min})$ to $(r_\mathrm{max}, c_\mathrm{max})$ in patch indices. The offset for each patch $p$ in $\mathcal{R}_\mathrm{low}$ is defined as:
\begin{equation}
\Delta r = r - r_\mathrm{min}, \quad \Delta c = c - c_\mathrm{min}.\label{eq:delta}
\end{equation}

To replace the low-confidence region, we search for a matching region in another confidence map with the same shape as $\mathcal{R}_\mathrm{low}$. For each possible top-left corner $(r_s, c_s)$ in the second map, we define a candidate region $\mathcal{R}_c$ with the same offsets as $\mathcal{R}_\mathrm{low}$. The average confidence of $\mathcal{R}_c$ is computed as:
\begin{equation}
\overline{C}(\mathcal{R}_c) = \frac{1}{|\mathcal{R}_c|} \sum_{p \in \mathcal{R}_c} \widetilde{C}_{p},\label{eq:crc}
\end{equation}
where $|\mathcal{R}_c| = |\mathcal{R}_\mathrm{low}|$.

We identify the candidate region with the highest average confidence:
\begin{equation}
\mathcal{R}_\mathrm{high} = \arg\max_{\mathcal{R}_c} \overline{C}(\mathcal{R}_c).\label{eq:Rhigh}
\end{equation}

Finally, the patches in $\mathcal{R}_\mathrm{low}$ are replaced with the corresponding patches from $\mathcal{R}_\mathrm{high}$ in the input image, ensuring that the region with the lowest confidence is updated with the most reliable information from the other model.

\subsubsection{Dynamic Threshold Escalation (DTE)}

To adaptively guide the training process, we introduce two thresholds that evolve dynamically during training: the confidence threshold \( C_\mathrm{threshold}(t) \) and the maximum region size \( R_\mathrm{threshold}(t) \). These thresholds are defined as follows:
\begin{equation}
\begin{aligned}
C_\mathrm{threshold}(t) 
&= 
C_\mathrm{min} 
+ 
\big(C_\mathrm{max} - C_\mathrm{min}\big) \Psi(t), \\
R_\mathrm{threshold}(t) 
&= 
R_\mathrm{min} 
+ 
\big(R_\mathrm{max} - R_\mathrm{min}\big) \Psi(t),
\end{aligned}
\label{eq:thresholds}
\end{equation}

where \( \Psi(t) \) is a monotonically increasing ramp function that maps the training iteration \( t \) to a value between 0 and 1:
\begin{equation}
\Psi(t) = 1 - e^{-t / \beta}. \label{eq:psi}
\end{equation}

This strategy is motivated by the distinct requirements of different training stages. During the early stages, the model's predictions are highly uncertain and less reliable. To ensure stable learning, we set \( C_\mathrm{threshold}(t) \) to a low value (\( C_\mathrm{min} \)), restricting the replacement process to only the least confident regions. Similarly, the size of the connected regions is limited by setting \( R_\mathrm{threshold}(t) \) to \( R_\mathrm{min} \), preventing extensive modifications that could destabilize the model. As training progresses and the model becomes more confident, both thresholds increase, allowing the replacement process to encompass a larger portion of the image domain. By the final stages of training, the thresholds reach \( C_\mathrm{max} \) and \( R_\mathrm{max} \), enabling comprehensive and aggressive refinement of the segmentation predictions.

\subsection{Loss Function}

The loss function is composed of three main components: Mean Teacher Loss (\(L_\text{MT}\)), CPS Loss (\(L_\text{CPS}\)), and CAD Loss (\(L_\text{CAD}\)). Each of these losses is designed to guide the model's training, using a combination of both labeled and unlabeled data to enhance the segmentation performance.

\subsubsection{Mean Teacher Loss (\(L_\text{MT}\))}

The Mean Teacher loss is calculated using both labeled and unlabeled data, employing a combination of Dice loss and Cross Entropy loss. The Mean Teacher loss is computed separately for weakly augmented data (\(L_\text{MT1}\)) and strongly augmented data (\(L_\text{MT2}\)).

For labeled data, the Mean Teacher loss combines the Dice loss and the Cross Entropy loss, given by:
\begin{equation}
L_{\text{MT}} = \left[ 1 - \frac{2 \sum_{c}\text{true}_c \cdot \hat{\text{pred}}_c}{\sum_{c}\text{true}_c + \sum_{c}\hat{\text{pred}}_c} \right] - \sum_{c}\text{true}_c \log(\hat{\text{pred}}_c),
\end{equation}
where \(\text{true}_c\) and \(\hat{\text{pred}}_c = \text{softmax}(\text{logits}_c)\) represent the ground truth and the predicted probability for class \(c\), respectively.

For unlabeled data, the loss function is similarly computed, but only the pseudo labels generated by the model are used for both the Dice and Cross Entropy losses. We denote the loss for unlabeled weakly augmented data as \(L_{\text{MT1}}\) and for strongly augmented data as \(L_{\text{MT2}}\).

Thus, the total Mean Teacher Loss is the sum of the weak and strong augmentation losses:
\begin{equation}
L_\text{MT1} = L_{\text{dice}}(f_{\theta_1}, p) + L_{\text{ce}}(f_{\theta_1}, p), 
\end{equation}
\begin{equation}
L_\text{MT2} = L_{\text{dice}}(f_{\theta_2}, p) + L_{\text{ce}}(f_{\theta_2}, p).
\end{equation}

\subsubsection{Cross Pseudo Supervision Loss (\(L_\text{CPS}\))}

The Cross Pseudo Supervision Loss ensures that the predictions from two different student models are consistent with each other. For weakly augmented data, we define the CPS loss as \(L_{\text{CPS1}}\), and for strongly augmented data, we define it as \(L_{\text{CPS2}}\).

The CPS loss is computed as the Dice loss between the softmax outputs of the two models, which are \(f_{\theta_1}\) and \(f_{\theta_2}\), using the pseudo labels generated by each model for the other. For weakly augmented data:
\begin{equation}
L_{\text{CPS1}} = \text{dice\_loss}(\text{softmax}(f_{\theta_1}(X_w)), \text{argmax}(f_{\theta_2}(X_s))),
\end{equation}
and for strongly augmented data:
\begin{equation}
L_{\text{CPS2}} = \text{dice\_loss}(\text{softmax}(f_{\theta_2}(X_s)), \text{argmax}(f_{\theta_1}(X_w))).
\end{equation}
The Dice loss is calculated similarly to the equation for \(L_{\text{dice}}\), where the predicted labels are compared with the pseudo labels generated by the opposite student model.

\subsubsection{Confidence-Aware Displacement Loss (\(L_\text{CAD}\))}

The CAD loss is computed after applying CAD to replace the low-confidence regions in the image. Following the replacement, the modified images are passed through the models, and Cross Pseudo Supervision loss is calculated based on the models' predictions. The CAD loss is computed for both weakly augmented data and strongly augmented data, denoted as \(L_{\text{CAD1}}\) and \(L_{\text{CAD2}}\), respectively.

After the low-confidence regions are replaced, we denote the new image after replacement as \(X'_w\) for the weakly augmented data and \(X'_s\) for the strongly augmented data, where these images are now the modified versions, containing the high-confidence regions from the other model's predictions.

For weakly augmented data after replacement, the CAD loss is computed as:
\begin{equation}
L_{\text{CAD1}} = \text{dice\_loss}(\text{softmax}(f_{\theta_1}(X'_w)), \text{argmax}(f_{\theta_2}(X'_s))),
\end{equation}
where \(X'_w\) represents the weakly augmented image after low-confidence regions have been replaced with high-confidence patches from the strongly augmented image \(X'_s\).

For strongly augmented data after replacement, the CAD loss is given by:
\begin{equation}
L_{\text{CAD2}} = \text{dice\_loss}(\text{softmax}(f_{\theta_2}(X'_s)), \text{argmax}(f_{\theta_1}(X'_w))).
\end{equation}
Here, \(X'_s\) represents the strongly augmented image after low-confidence regions have been replaced with high-confidence patches from the weakly augmented image \(X'_w\).

The CAD loss functions \(L_{\text{CAD1}}\) and \(L_{\text{CAD2}}\) ensure that the replaced low-confidence regions are consistent with the rest of the image and that the pseudo labels produced by the other model are in agreement with the modifications made to the image.


\subsubsection{Total Loss}

The total loss for the model is computed by summing the Mean Teacher Loss, Cross Pseudo Supervision Loss, and CAD Loss for both weak and strong augmented data. The total loss for the first model is given by:
\begin{equation}
L_1 = L_{\text{MT1}} + L_{\text{CPS1}} + L_{\text{CAD1}},
\end{equation}
and the total loss for the second model is given by:
\begin{equation}
L_2 = L_{\text{MT2}} + L_{\text{CPS2}} + L_{\text{CAD2}}.
\end{equation}

The final total loss is the sum of \(L_1\) and \(L_2\):
\begin{equation}
L = L_1 + L_2.
\end{equation}

This loss function drives the network to learn accurate segmentation by leveraging both labeled and unlabeled data, dynamically guiding the model to focus on uncertain regions and progressively refining its predictions through a combination of each loss components.

\section{Experiments}

\subsection{Datasets}

In this study, we evaluate the proposed semi-supervised learning approach using two widely recognized datasets: ACDC (Automated Cardiac Diagnosis Challenge)\cite{b17} and PROMISE12\cite{b18}. Although originally composed of 3D medical images, both datasets are converted into 2D slices for simplicity, aligning with common practices in segmentation studies\cite{b19}.

\subsubsection{ACDC Dataset} The ACDC dataset\cite{b17} contains 200 short-axis cardiac cine-MR images from 100 patients, labeled into four classes: left ventricle, right ventricle, myocardium, and background. Each 3D scan is converted into 2D cross-sectional slices, and the dataset is partitioned into training, validation, and testing sets with a ratio of 70:10:20.

\subsubsection{PROMISE12 Dataset} The PROMISE12 dataset\cite{b18}, designed for the MICCAI 2012 prostate segmentation challenge, includes MRI scans from 50 patients with diverse prostate pathologies. We similarly convert the original 3D images into 2D slices, which pose segmentation challenges due to substantial variations in prostate size and shape among patients.

\subsection{Evaluation Metrics}

For the evaluation of our model, we use four commonly used metrics in medical image segmentation: Dice Similarity Coefficient (DSC), Jaccard Index (Jaccard), 95\% Hausdorff Distance (95HD), and Average Surface Distance (ASD). 

The DSC and Jaccard Index are both used to measure the overlap between the predicted and ground truth regions, with higher values indicating better segmentation performance. 

The ASD computes the average distance between the boundaries of the predicted and ground truth regions, giving an indication of the accuracy of boundary localization. 

The 95HD measures the maximum distance between the boundaries of the predicted and ground truth regions, considering the 95th percentile, and provides insight into the largest discrepancy between the two regions.

\subsection{Implementation Details}

We conducted our experiments in an NVIDIA V100 GPU environment. The image patch is divided into \(N_p\) smaller patches, where each patch is of size \(h \times w\). In our experiments, we set \(h = w = 16\), leading to a total of \(N_p = 256\) patches for each image.

The confidence threshold \(C_\text{threshold}(t)\) and the region size threshold \(R_\text{threshold}(t)\) evolve during the training process. Initially, the confidence threshold is set to a small value (\(C_\text{min} = 0.01\)) and the region size threshold is set to a small value (\(R_\text{min} = 1\)). These values gradually increase over training, with the confidence threshold reaching a maximum of \(C_\text{max} = 0.75\) and the region size threshold growing to \(R_\text{max} = 16\) by the end of the training process.

Our method employs the Mean Teacher framework, which is built upon the BCP\cite{b3} strategy to enhance performance.

\subsection{Comparison with State-of-the-Art Methods} 
\subsubsection{ACDC Dataset}

We compare our proposed CAD method with ABD\cite{b5} (the baseline method), CorrMatch\cite{b28}, SAMT-PCL\cite{b29}, BCP\cite{b3}, SCP-Net\cite{b6}, and other methods on the ACDC test set using 5\% and 10\% of labeled data for training. The results are presented in Table.~\ref{tab:acdc_comparison}, where the metrics of ABD and CorrMatch are reproduced. As shown, CAD outperforms the baseline model when trained with 5\% labeled data, achieving higher scores in DSC, 95HD, and ASD.

\begin{figure}[hbtp]
\centerline{\includegraphics[width=250pt]{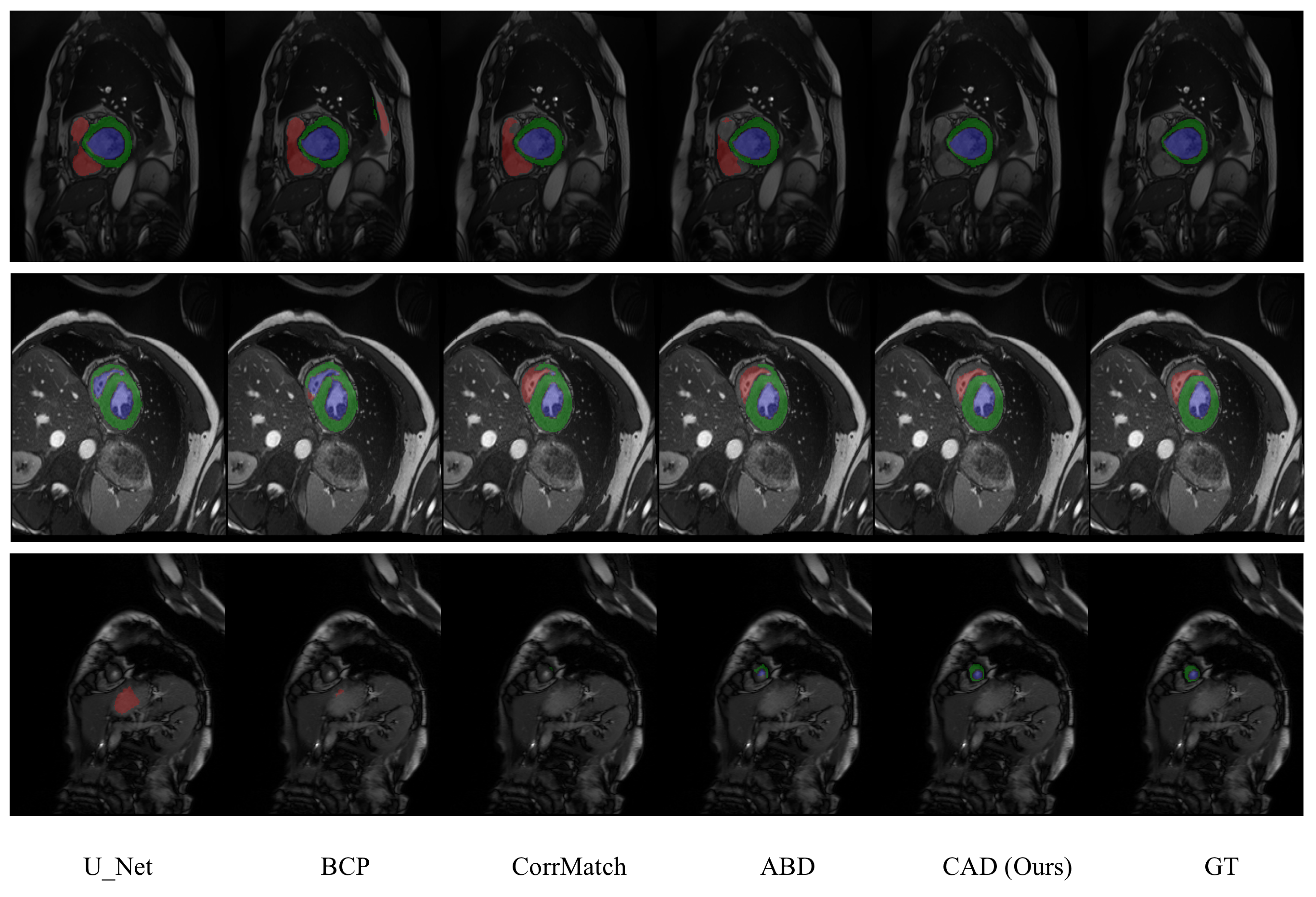}}
\caption{Segmentation results on three test set samples from the ACDC dataset using 10\% labeled data: U-Net, BCP, CorrMatch, ABD, CAD (Ours), and Ground Truth (GT).}
\label{fig3}
\end{figure}

When trained with 10\% labeled data, our model demonstrates superior performance across all four evaluation metrics compared to existing state-of-the-art methods. Notably, CAD achieves a remarkable 42.7\% improvement in 95HD, setting a new benchmark for state-of-the-art performance on this dataset. The comparison of three samples from the test set with U-Net\cite{b22}, BCP\cite{b3}, CorrMatch\cite{b28}, ABD\cite{b5}, and Ground Truth, all trained with 10\% labeled data, are shown in Fig.~\ref{fig3}

\begin{table*}[htbp]
\caption{Comparisons with other state-of-the-art methods on the ACDC test set.}
\begin{center}
\begin{tabular}{|l|c|c|c|c|c|c|c|}
\hline
\textbf{Method} & \multicolumn{2}{|c|}{\textbf{Scans used}} & \multicolumn{4}{|c|}{\textbf{Metrics}} \\
\cline{2-7} 
 & Labeled & Unlabeled & DSC $\uparrow$ & Jaccard $\uparrow$ & 95HD $\downarrow$ & ASD $\downarrow$ \\
\hline
U-Net (MICCAI'2015) \cite{b22} & 3 (5\%) & 0 & 47.83 & 37.01 & 31.16 & 12.62 \\
 & 7 (10\%) & 0 & 79.41 & 68.11 & 9.35 & 2.70 \\
 & 70 (All) & 0 & 91.44 & 84.59 & 4.30 & 0.99 \\
\hline
DTC (AAAI'2021) \cite{b25} & 3 (5\%) & 67 (95\%) & 56.90 & 45.67 & 23.36 & 7.39 \\
URPC (MICCAI'2021) \cite{b24} & 3 (5\%) & 67 (95\%) & 55.87 & 44.64 & 13.60 & 3.74 \\
MC-Net (MICCAI'2021) \cite{b23} & 3 (5\%) & 67 (95\%) & 62.85 & 52.29 & 7.62 & 2.33 \\
SS-Net (MICCAI'2022) \cite{b21} & 3 (5\%) & 67 (95\%) & 65.83 & 55.38 & 6.67 & 2.28 \\
SCP-Net (MICCAI'2023) \cite{b6} & 3 (5\%) & 67 (95\%) & 87.27 & - & - & 2.65 \\
BCP (CVPR'2023) \cite{b3} & 3 (5\%) & 67 (95\%) & 87.59 & 78.67 & 1.90 & 0.67 \\
SAMT-PCL (ESA'2024) \cite{b29} & 3 (5\%) & 67 (95\%) & 74.39 & 63.94 & 5.07 & 1.42 \\
CorrMatch (CVPR'2024) \cite{b28} & 3 (5\%) & 67 (95\%) & 86.33 & 75.99 & 4.01 & 1.22 \\
ABD (CVPR'2024) \cite{b5} & 3 (5\%) & 67 (95\%) & 87.98 & \textbf{80.12} & 1.89 & 0.74 \\
\textbf{CAD (Ours)} & 3 (5\%) & 67 (95\%) & \textbf{88.02} & 79.79 & \textbf{1.55} & \textbf{0.48} \\
\hline
DTC (AAAI'2021) \cite{b25} & 7 (10\%) & 63 (90\%) & 84.29 & 73.92 & 12.81 & 4.01 \\
URPC (MICCAI'2021) \cite{b24} & 7 (10\%) & 63 (90\%) & 83.10 & 72.41 & 4.84 & 1.53 \\
MC-Net (MICCAI'2021) \cite{b23} & 7 (10\%) & 63 (90\%) & 86.44 & 77.04 & 5.50 & 1.84 \\
SS-Net (MICCAI'2022) \cite{b21} & 7 (10\%) & 63 (90\%) & 86.78 & 77.67 & 6.07 & 1.40 \\
SCP-Net (MICCAI'2023) \cite{b6} & 7 (10\%) & 63 (90\%) & 89.69 & - & - & 0.73 \\
PLGCL (CVPR'2023) \cite{b20} & 7 (10\%) & 63 (90\%) & 89.1 & - & 4.98 & 1.80 \\
BCP (CVPR'2023) \cite{b3} & 7 (10\%) & 63 (90\%) & 88.84 & 80.62 & 3.98 & 1.17 \\
SAMT-PCL (ESA'2024) \cite{b29} & 7 (10\%) & 63 (90\%) & 88.62 & 80.11 & 2.06 & 0.60 \\
CorrMatch (CVPR'2024) \cite{b28} & 7 (10\%) & 63 (90\%) & 87.33 & 78.27 & 4.31 & 1.45 \\
ABD (CVPR'2024) \cite{b5} & 7 (10\%) & 63 (90\%) & 89.77 & 81.88 & 1.77 & 0.49 \\
\textbf{CAD (Ours)}  & 7 (10\%) & 63 (90\%) & \textbf{90.38} & \textbf{82.89} & \textbf{1.24} & \textbf{0.36} \\
\hline
\end{tabular}
\label{tab:acdc_comparison}
\end{center}
\end{table*}

\subsubsection{PROMISE12 Dataset}

On the PROMISE12 dataset, we compare our proposed CAD method with ABD\cite{b5} (the baseline model), CorrMatch\cite{b28}, BCP\cite{b3}, SCP-Net\cite{b6}, SLC-Net\cite{b27}, SS-Net\cite{b21}, URPC\cite{b24}, and CCT\cite{b26} using 20\% of labeled data for training. The comparison focuses on the DSC and ASD metrics. As shown in Table.~\ref{tab:promise12_comparison}, the results for ABD, BCP, and CorrMatch are based on our reproduced data. The experimental outcomes demonstrate that CAD outperforms the baseline model in both metrics and achieves state-of-the-art performance in terms of DSC.

\begin{table}[htbp]
\caption{Comparisons with state-of-the-art methods on the PROMISE12 test set.}
\begin{center}
\resizebox{0.5\textwidth}{!}{
\begin{tabular}{|l|c|c|c|c|}
\hline
\textbf{Method} & \multicolumn{2}{|c|}{\textbf{Scans used}} & \multicolumn{2}{|c|}{\textbf{Metrics}} \\
\cline{2-5} 
 & Labeled & Unlabeled & DSC $\uparrow$ & ASD $\downarrow$ \\
\hline
U-Net (MICCAI'2015) \cite{b22} & 7 (20\%) & 0 & 60.88 & 13.87 \\
 & 35 (100\%) & 0 & 84.76 & 1.58 \\
\hline
CCT (CVPR'2020) \cite{b26} & 7 (20\%) & 28 (80\%) & 71.43 & 16.61 \\
URPC (MICCAI'2021) \cite{b24} & 7 (20\%) & 28 (80\%) & 63.23 & 4.33 \\
SS-Net (MICCAI'2022) \cite{b21} & 7 (20\%) & 28 (80\%) & 62.31 & 4.36 \\
SLC-Net (MICCAI'2022) \cite{b27} & 7 (20\%) & 28 (80\%) & 68.31 & 4.69 \\
SCP-Net (MICCAI'2023) \cite{b6} & 7 (20\%) & 28 (80\%) & 77.06 & 3.52 \\
BCP (CVPR'2023) \cite{b3} & 7 (20\%) & 28 (80\%) & 75.54 & 2.88 \\
SAMT-PCL (ESA'2024) \cite{b29} & 7 (20\%) & 28 (80\%) & 78.35 & \textbf{1.81} \\
CorrMatch (CVPR'2024) \cite{b28} & 7 (20\%) & 28 (80\%) & 79.77 & 4.31 \\
ABD (CVPR'2024) \cite{b5} & 7 (20\%) & 28 (80\%) & 77.52 & 2.23 \\
\textbf{CAD (Ours)} & 7 (20\%) & 28 (80\%) & \textbf{79.80} & 2.01 \\
\hline
\end{tabular}
}
\label{tab:promise12_comparison}
\end{center}
\end{table}

\subsection{Ablation Studies}





\subsubsection{Effectiveness of Each Module in CAD}
To validate the effectiveness of each component, we conducted an ablation study on the ACDC dataset using 10\% of labeled data, as shown in Table.~\ref{tab:effectiveness}. In Table.~\ref{tab:effectiveness}, the first column, \textbf{BL}, represents the baseline model, which does not apply any replacement strategy. The second column, \textbf{LLCR}, and the third column, \textbf{DTE}, denote the Largest Low-Confidence Region Replacement and Dynamic Threshold Escalation strategies, respectively.

In the ABD\cite{b5}, for unlabeled data, the replacement process first calculates the Kullback-Leibler (KL) divergence\cite{b30} between the top \(K_\text{top}\) high-confidence patches and the candidate low-confidence patches. The patch with the minimum KL divergence, representing the closest output distribution, is then selected for replacement. Following a similar approach, we performed an ablation analysis in our study. After identifying the low-confidence region, we calculated its KL divergence with the logits of the \(K_\text{top}\) high-confidence regions as follows:

\begin{equation}
\mathrm{KL}(\mathbf{z}_\mathrm{low}, \mathbf{z}^\mathrm{i}_{high}) = \sum_c \mathrm{softmax}(\mathbf{z}_\mathrm{low})_c \log \frac{\mathrm{softmax}(\mathbf{z}_\mathrm{low})_c}{\mathrm{softmax}(\mathbf{z}^\mathrm{i}_{high})_c},
\end{equation}

where \(\mathbf{z}_\mathrm{low}\) and \(\mathbf{z}^\mathrm{i}_{high}\) represent the logits of the low-confidence region and the \(i\)-th high-confidence region in \(K_\text{top}\), respectively. Finally, the high-confidence region with the smallest KL divergence is selected for replacement. This strategy is abbreviated as \textbf{KL} in the fourth column of Table.~\ref{tab:effectiveness}.

\begin{table}[htbp]
\caption{Ablation studies on the ACDC dataset with 10\% labeled data, validating the effectiveness of each component.}
\begin{center}
\begin{tabular}{|c c c c|c c c c|}
\hline
\textbf{BL} & \textbf{LLCR} & \textbf{DTE} & \textbf{KL} & \textbf{DSC$\uparrow$} & \textbf{Jaccard$\uparrow$} & \textbf{95HD$\downarrow$} & \textbf{ASD$\downarrow$} \\
\hline
\checkmark &  &  &  & 89.03 & 80.79 & 3.34 & 0.94 \\
\checkmark & \checkmark &  &  & 89.23 & 81.69 & 2.44 & 0.82 \\
\checkmark & \checkmark & \checkmark &  & \textbf{90.38} & \textbf{82.89} & \textbf{1.29} & \textbf{0.36} \\
\checkmark & \checkmark & \checkmark & \checkmark & 89.69 & 81.78 & 1.82 & 0.69 \\

\hline
\end{tabular}
\label{tab:effectiveness}
\end{center}
\end{table}

The results demonstrate that the proposed CAD method benefits significantly from the inclusion of both the LLCR and DTE strategies, which play a positive role in enhancing the model's learning capability under strong perturbations. In contrast, the KL strategy does not show a notable positive impact on model performance within the CAD framework. Therefore, this component is not incorporated into our final CAD approach.

\subsubsection{Selection of Threshold Hyperparameters in DTE}
Table.~\ref{tab:hyperparameters} illustrates the impact of different choices for the hyperparameters \(C_\text{min}\), \(C_\text{max}\), \(R_\text{min}\), and \(R_\text{max}\) on the model's performance within the DTE strategy. These hyperparameters define the minimum and maximum confidence thresholds for patches to be included in the low-confidence connected region \(\mathcal{R}_\mathrm{low}\), as well as the minimum and maximum allowable number of patches in \(\mathcal{R}_\mathrm{low}\). As shown in the table, the model achieves the best performance when \(C_\text{min}\), \(C_\text{max}\), \(R_\text{min}\), and \(R_\text{max}\) are set to 0.01, 0.75, 1, and 16, respectively.

\begin{table}[htbp]
\caption{Ablation Study of Threshold Hyperparameters on ACDC Dataset with 10\% Labeled Data.}
\begin{center}
\resizebox{0.5\textwidth}{!}{
\begin{tabular}{|c c c c|c c c c|}
\hline
\textbf{$C_\text{min}$} & \textbf{$C_\text{max}$} & \textbf{$R_\text{min}$} & \textbf{$R_\text{max}$} & \textbf{DSC$\uparrow$} & \textbf{Jaccard$\uparrow$} & \textbf{95HD$\downarrow$} & \textbf{ASD$\downarrow$} \\
\hline
0.1 & 0.9 & 4 & 32 & 88.22 & 79.76 & 4.77 & 3.00 \\
0.05 & 0.9 & 4 & 32 & 88.34 & 80.01 & 3.97 & 3.21 \\
0.01 & 0.9 & 4 & 32 & 88.43 & 79.89 & 2.44 & 2.56 \\
0.01 & 0.6 & 4 & 32 & 88.27 & 80.29 & 2.37 & 1.74 \\
0.01 & 0.75 & 4 & 32 & 89.14 & 81.21 & 1.76 & 0.87 \\
0.01 & 0.75 & 8 & 32 & 88.55 & 79.79 & 3.24 & 2.94 \\
0.01 & 0.75 & 1 & 32 & 89.69 & 81.79 & 1.97 & 0.51 \\
0.01 & 0.75 & 1 & 8 & 89.83 & 81.83 & 1.44 & \textbf{0.34} \\

0.01 & 0.75 & 1 & 16 & \textbf{90.38} & \textbf{82.89} & \textbf{1.29} & 0.36 \\

\hline
\end{tabular}
}
\label{tab:hyperparameters}
\end{center}
\end{table}

\section{Conclusion}
In this work, we proposed the Confidence-Aware Displacement (CAD) strategy for semi-supervised medical image segmentation, incorporating the Largest Low-Confidence Region Replacement and Dynamic Threshold Escalation strategies. By adaptively identifying and replacing low-confidence regions, our approach effectively enhances the model's learning from unlabeled data while ensuring stability during training. Extensive experiments on the ACDC and PROMISE12 datasets demonstrate the superiority of CAD, achieving state-of-the-art performance with limited labeled data and outperforming other methods. These results highlight the effectiveness of CAD in using uncertainty to refine segmentation predictions, providing a robust framework for improving semi-supervised segmentation tasks in medical imaging.

\bibliography{CAD.bib}
\vspace{12pt}

\end{document}